\documentclass[twoside,11pt]{article}

\usepackage[preprint]{dmlr2e}
\usepackage{dmlr2e}

\usepackage{hyperref}
\usepackage{url}

\usepackage{tikz}
\usepackage{pgfplots}
\usepackage{subcaption}
\usepackage{multirow}
\usepackage{dsfont}
\usepackage{makecell}
\usepackage{booktabs}  
\usepackage{nicematrix}
\usepackage{rotating}
\usepackage{amsfonts}       
\usepackage{nicefrac}       
\usepackage{microtype}      
\usepackage{xcolor}         
\usepackage{xspace}
\usepackage{wrapfig}

\usepackage{xcolor}
\hypersetup{
    colorlinks,
    linkcolor={black!50!black},
    citecolor={black!50!black},
    urlcolor={black!80!black}
}

\usepackage{enumitem}
\usepackage{longtable, lmodern}
\usepackage{caption}
\newcommand{\mpara}[1]{\medskip\noindent{\bf #1}}
\newcommand{\blog}{\textsc{BlogCat}\xspace}
\newcommand{\yelp}{\textsc{Yelp}\xspace}
\newcommand{\pcg}{\textsc{PCG}\xspace}

\newcommand{\ogb}{\textsc{OGB-Proteins}\xspace}
\newcommand{\dblp}{\textsc{DBLP}\xspace}

\newcommand{\humloc}{\textsc{HumLoc}\xspace}
\newcommand{\eukloc}{\textsc{EukLoc}\xspace}
\newcommand{\gcn}{\textsc{Gcn}\xspace}
\newcommand{\mlp}{\textsc{Mlp}\xspace}
\newcommand{\gat}{\textsc{Gat}\xspace}
\newcommand{\deepwalk}{\textsc{DeepWalk}\xspace}
\newcommand{\graphsage}{\textsc{GraphSage}\xspace}
\newcommand{\hgcn}{\textsc{H2Gcn}\xspace}
\newcommand{\gcnlpa}{\textsc{GCN-LPA}\xspace}
\newcommand{\lanc}{\textsc{LANC}\xspace}
\newcommand{\openreview}{} 

\usepackage{lastpage}
\dmlrheading{24}{2024}{1-\pageref{LastPage}}{06/24; Revised 06/24}{11/23}{21-0000}{Tianqi Zhao, Ngan Thi Dong, Alan Hanjalic, Megha Khosla} 
\def\openreview{\url{https://openreview.net/forum?id=rnAxUZCHor}} 

\ShortHeadings{A data-centric approach for assessing progress of Graph Neural Networks}{Zhao, Dong, Hanjalic, Khosla}
\firstpageno{1}

\title{A data-centric approach for assessing progress of Graph Neural Networks }

\author{\name Tianqi Zhao \email T.Zhao-1@tudelft.nl \\
       \addr Department of Intelligent Systems\\
       Delft University of Technology \\
       Delft, the Netherlands
       \AND
       \name Ngan Thi Dong \email dong@l3s.de \\
      \addr L3S Research Center\\
      Hannover, Germany\\
       \AND
       \name Alan Hanjalic \email A.Hanjalic@tudelft.nl \\
       \addr Division of Computer Science\\
       Delft University of Technology \\
       Delft, the Netherlands
       \AND
       \name Megha Khosla \email M.Khosla@tudelft.nl \\
       \addr Division of Computer Science\\
       Delft University of Technology \\
       Delft, the Netherlands}

\begin{document}

\maketitle

\section{Introduction}
Graph Neural Networks (GNNs) have shown state-of-the-art improvements in node classification tasks on graphs. While these improvements have been largely demonstrated in a multi-class classification scenario, a more general and realistic scenario in which each node could have multiple labels has so far received little attention. In this work we provide a data-centric analysis of the performance of GNNs on multi-label datasets. 

We start by analysing the existing datasets for multi-label node classification which already pointed to severe data quality issues. As an example, the sole multi-label node classification dataset in the Open Graph Benchmark (OGB) \citep{hu2020ogb}, namely \ogb, has around $90$\% of the nodes unlabeled in the test set. While the OGB leaderboard reflects benchmarking of a large number of methods on \ogb, the lack of labels in the test set combined with the use of the Area Under the ROC Curve (AUROC) metric leads to overly exaggerated performance scores. 

Furthermore, the success of GNNs is largely attributed to feature smoothing over neighborhoods and high label similarity among neighboring nodes, known as homophily. Conversely, heterophilic graphs have neighboring nodes with dissimilar labels. However, our analyses show that multi-label networks do not fit neatly into these categories. For example, in a social network, users might share only a few interests with their friends, indicating low local homophily, yet their interests could still be inferred from their neighbors, pointing to the fact that nodes characterized by multiple labels does not obey the crisp separation of homophilic and heterophilic characteristics as studied so far for nodes with a single label. As a result, methods developed and tested for multi-class datasets taking such characteristics cannot be directly deployed for  multi-label node classification.

To assess the actual progress made by current graph machine learning approaches for multi-label node classification we develop a three-dimensional data-centric strategy in which we (i) investigate characteristics of multi-label graph datasets, such as label distribution and label-induced similarities and analyse their influence on model performance, (ii) curate three biological graph datasets and develop a synthetic multi-label graph generator with tunable properties, enabling rigorous comparison of learning methods, and (iii) conduct a large-scale experimental study evaluating eight methods across nine datasets. Simple baselines like DeepWalk outperform more sophisticated GNNs on several datasets. 

\paragraph{Remark:}This paper is an extended abstract corresponding to the published paper \citet{zhao2023multilabel}. The corresponding code is available at: \url{https://github.com/Tianqi-py/MLGNC}.
\section{Problem Setting and Analysis of Datasets}
\label{sec:notations}

Let $\mathcal{G} = (\mathcal{V}, \mathcal{E})$ denote a graph where $\mathcal{V}=\left\{v_{1}, \cdots, v_{n}\right\}$ is the set of vertices, $\mathcal{E}$ represents the set of links/edges among the vertices. We further denote the adjacency matrix of the graph by $\mathbf{A} \in\{0,1\}^{n \times n}$ and $a_{i,j}$ denotes whether there is an edge between $v_{i}$ and $v_{j}$. $\mathcal{N}(v)$ represents the immediate neighbors of node $v$ in the graph.
Furthermore, let $\mathbf{X}=\left\{\mathbf{x}_{1}, \cdots, \mathbf{x}_{n}\right\} \in \mathbb{R}^{n \times D}$ and $\mathbf{Y}=\left\{\mathbf{y}_{1}, \cdots, \mathbf{y}_{n}\right\} \in \{0,1\}^{n \times C}$ represent the feature and label matrices corresponding to the nodes in $\mathcal{V}$. In the feature matrix and label matrix, the  $i$-th row represents the feature/label vector of node $i$. Let $\ell(i)$ denote the set of labels that are assigned to node $i$. 
Finally, let $\mathcal{F}$ correspond to the feature set and $\mathcal{L}$ be the set of all labels. We are given a set of labeled $\mathcal{V}_\ell \subset \mathcal{V}$ and unlabelled nodes $ \mathcal{V}\setminus \mathcal{V}_\ell$ where the complete label set of each of the labelled nodes is known . We are then interested in predicting labels of unlabelled nodes. We assume that the training nodes are completely labeled. We deal with the transductive setting multi-label node classification problem, where the features and graph structure of the test nodes are present during training. 

\paragraph{Dataset Analysis.} We begin by analyzing various properties of existing multi-label datasets, such as label distributions, label similarities, and cross-class neighborhood similarity (CCNS) which are defined below.
\begin{definition}
\label{def:lable_homo}
Given a multi-label graph $\mathcal{G}$, the label homophily $h$ of $\mathcal{G}$ is defined as the average of the Jaccard similarity of the label set of all connected nodes in the graph:
$
    h = {1\over  |\mathcal{E}|}\sum_{(i,j)\in \mathcal{E}} {{|\ell(i) \cap \ell(j)|\over  |\ell(i) \cup \ell(j)|} }
$
\end{definition}

\begin{definition}
 Given a multi-label graph $\mathcal{G}$ and the set of node labels \textbf{$Y$} for all nodes, we define the multi-label cross-class neighborhood similarity between classes $c, c' \in C$ is given by $
 s(c, c')=\frac{1}{\left|\mathcal{V}_c \| \mathcal{V}_{c^{\prime}}\right|} \sum_{i \in \mathcal{V}_c, j \in \mathcal{V}_{c^{\prime}}, i \neq j} \frac{1}{|\ell(i)||\ell(j)|}\cos (\mathbf{d}_i, \mathbf{d}_j),$ where $\mathcal{V}_c =\{i|c\in \ell(i)\}$ is the set of nodes with one of their labels as $c$. The vector $\mathbf{d}_i \in \mathbb{R}^C$ corresponds to the empirical histogram (over $|C|$ classes) of node $i$'s neighbors' labels, i.e., the $c^{th}$ entry of $\mathbf{d}_i$ corresponds to the number of nodes in $\mathcal{N}(i)$ that has one of their label as $c$ and the function $cos(.,.)$ measures the cosine similarity.
\end{definition}

 \mpara{Analysis of existing datasets} We analyse  $4$ popular multi-label node classification datasets: (i) \blog \citep{Blogcatalog}, where nodes represent bloggers and edges their relationships, with labels denoting social groups, (ii) \yelp \citep{DBLP:journals/corr/abs-1907-04931}, where nodes correspond to customer reviews and edges to friendships, with labels representing business types, (iii) \ogb \citep{hu2020ogb}, where nodes represent proteins and edges indicate biologically meaningful associations, with labels corresponding to protein functions, and (iv) \dblp \citep{DBLP:journals/corr/abs-1910-09706}, where nodes represent authors, edges indicate co-authorship, and labels denote research areas. 

\paragraph{Skewed label distributions.} In \blog, $72.34\%$ of nodes have only one label, while the most labeled data points have 11 labels. \yelp has 100 labels, with the most labeled data points having 97 labels, and over $50\%$ of nodes having five or fewer labels. Despite this, \yelp demonstrates high multi-label character, with $75\%$ of nodes having more than three labels. In \ogb, $40.27\%$ of nodes lack any labels, while \dblp has the highest proportion of single-labeled nodes at $85.4\%$.

\paragraph{Issue in evaluation using AUROC scores under high label sparsity.} Another unreported issue in multi-label datasets is the presence of unlabeled data. In \ogb, $40.27\%$ of nodes lack labels, and $89.4\%$ of test nodes are unlabeled. Concerningly, the OGB leaderboard uses AUROC scores to benchmark methods for multi-label classification. However, a model that predicts "No Label" for each node (i.e., predicts the negative class for each binary classification task) achieves a high AUROC score. Increasing the number of training epochs, which encourages the model to predict the negative class, increases the AUROC score, while other metrics like AP or F1 score may drop or remain unchanged.

\begin{wrapfigure}{r}{0.5\textwidth}
     \centering
     \begin{subfigure}[b]{0.49\linewidth}
        \centering
         \includegraphics[width=\linewidth]{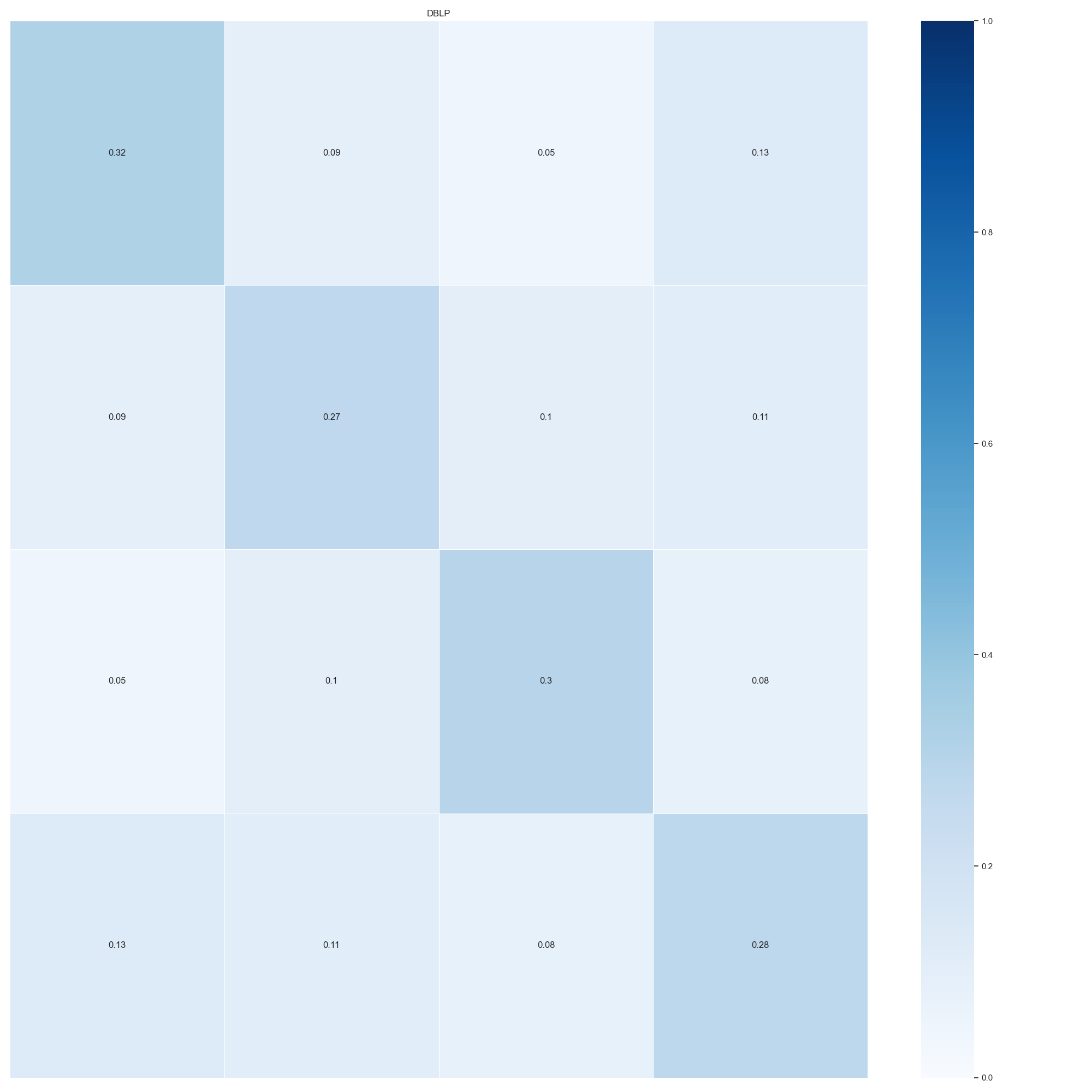}
         \caption{\dblp}
         \label{fig:ccns_dblp}
     \end{subfigure}
     \begin{subfigure}[b]{0.49\linewidth}
         \centering
       \includegraphics[width=\linewidth]{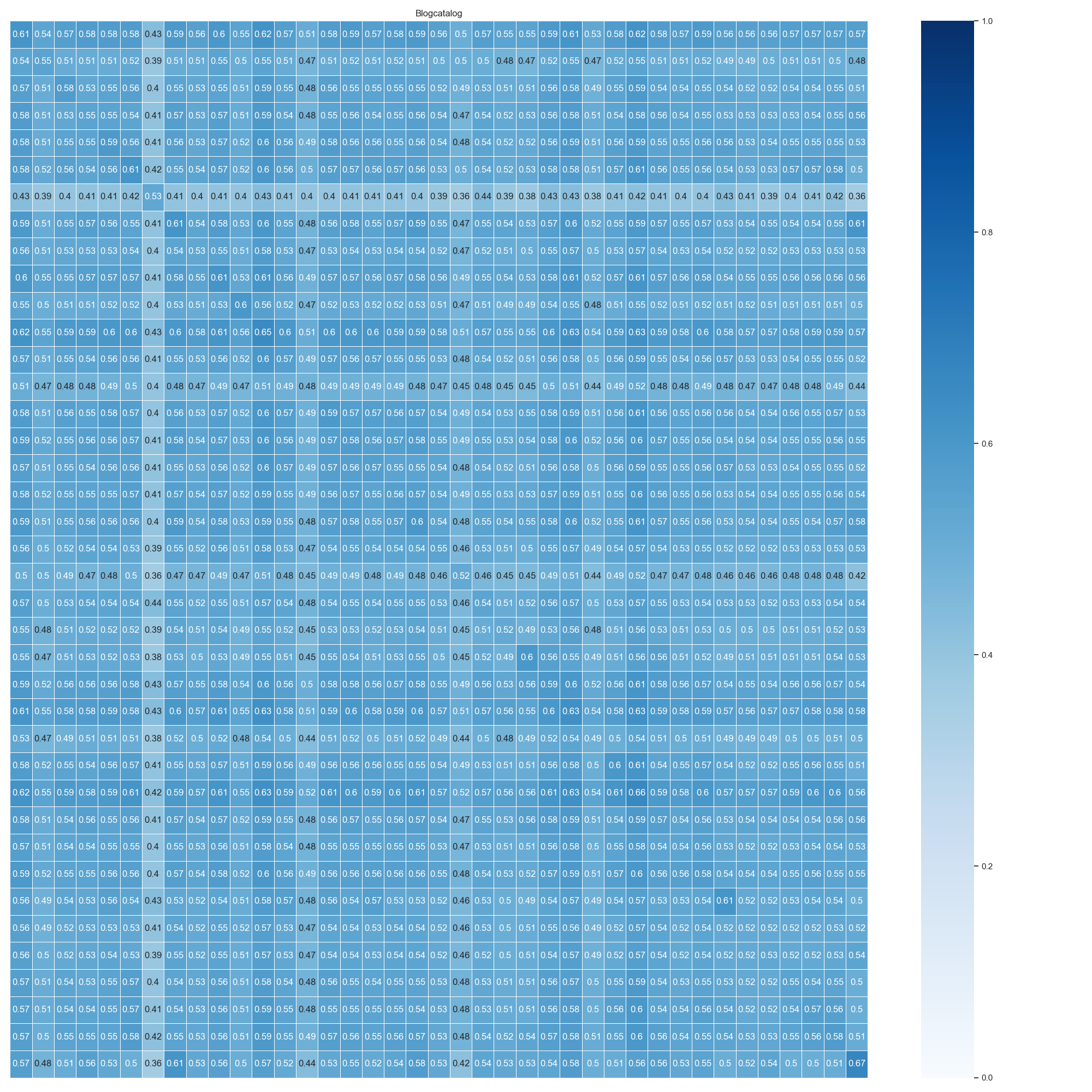}
         \caption{\blog}
         \label{fig:ccns_blogcatalog}
     \end{subfigure}
     \caption{Cross class Neighborhood Similarity in dataset \dblp and \blog.}
        \label{fig:ccns_real}
\end{wrapfigure}

\mpara{Cross-class neighborhood similarity.} In Figure \ref{fig:ccns_real}, we visualize the cross-class neighborhood similarity matrix for \dblp and \blog. The diagonal cells show intra-class neighborhood similarity, while others indicate inter-class similarity. The contrast in \ref{fig:ccns_dblp} suggests that nodes from the same class tend to have similar label distributions in their neighborhoods, aiding GNNs in correctly identifying nodes in the same classes in \dblp. Conversely, intra- and inter-class similarity are more similar in \blog, making it challenging for GNNs to classify nodes into their corresponding classes.

\section{Multi-label Graph Generator}
\label{sec:synthetic}

 To enable a comprehensive evaluation of the proposed models using multi-label datasets with diverse characteristics, we develop a multi-label graph generator model which facilitates   dataset construction with tunable properties, including high multi-label character, varying feature quality, label homophily, and CCNS similarity. Below, we describe the two main steps of our multi-label graph generator.

\textbf{Multi-label generator:} We generate a multi-label dataset using \textsc{Mldatagen} \citep{tomas2014framework}. First, we fix the total number of labels and features. We construct a hypersphere, $H \in \mathbb{R}^{|\mathcal{F}|}$, centered at the origin with a unit radius. For each label in set $\mathcal{L}$, we generate a smaller hypersphere contained within $H$. We populate these smaller hyperspheres with randomly generated data points of $|\mathcal{F}|$ dimensions. Each data point may lie in multiple overlapping hyperspheres, and its labels correspond to the hyperspheres it resides in.

\textbf{Graph generator:} Using the multi-label dataset, we construct edges between data points with a social distance attachment model \citep{boguna2004models}. For two data points (nodes) $i$ and $j$ with feature vectors $\mathbf{x}_i$ and $\mathbf{x}_j$ and label vectors $\mathbf{y}_i$ and $\mathbf{y}_j$, we denote the Hamming distance between their labels by $d(\mathbf{y}_i,\mathbf{y}_j)$. An edge between nodes $i$ and $j$ is constructed with probability:$ p_{ij} = \frac{1}{1 + [b^{-1}d(\mathbf{y}_i,\mathbf{y}_j)]^\alpha}$ where $\alpha$ is a homophily parameter, and $b$ is the characteristic distance at which $p_{ij} = \frac{1}{2}$. The parameters $\alpha$ and $b$ dictate edge density: a larger $b$ results in denser graphs, and a larger $\alpha$ assigns a higher connection probability to nodes with similar labels. This model, in large systems with high $\alpha$, leads to sparsity, non-trivial clustering, and positive degree assortativity, reflecting real-world network properties. By varying $\alpha$ and $b$, we control connection probability and label homophily. 

\paragraph{New biological datasets:} We propose $3$ biological datasets \pcg, \humloc, and \eukloc the details of which are in \citep{zhao2023multilabel}.

\section{Experiments}
\label{experiments}
We perform a large-scale empirical study comparing $8$ methods over $7$ real-world multi-label datasets and $2$ sets of synthetic datasets with varying homophily and feature quality. Our experimental results are provided in Tables \ref{tab: real_exres_ap} and \ref{tab: syn_exres}.   

\begin{table}[!h]
\setlength{\tabcolsep}{4pt}
\caption{Mean performance scores (Average Precision) on real-world datasets. }
\centering
\small
\begin{tabular}{l|ccccccc}
\hline
Method        &\blog            & \yelp             &\ogb     &\dblp &\pcg         &\humloc  &\eukloc \\ \hline
\mlp          &$0.043$      &$0.096$                 &$0.026$   &$0.350$ &$0.148$   &$0.170$  &$0.120$  \\ 
\deepwalk     &$\bf{0.190}$ &$0.096$                 &$0.044$   &$0.585$ &$\textbf{0.229}$    &$0.186$  &$0.076$    \\ 
\lanc         &$\underline{0.050}$ &OOM &$\underline{0.045}$ &$0.836$ &$0.185$ &$0.132$ &$0.062$  \\
\gcn          &$0.037$       &$0.131$                 &$\textbf{0.054}$   &$\bf{0.893}$ &$\underline{0.210}$   &$\textbf{0.252}$  &$\textbf{0.152}$ \\ 
\gat          &$0.041$       &$0.150$                 &$0.021$   &$0.829$ &$0.168$   &$\underline{0.238}$  &$\underline{0.136}$ \\ 
\graphsage    &$0.045$       &$\textbf{0.251}$     &$0.027$   &$\underline{0.868}$ &$0.185$   &$0.234$  &$0.124$\\
\hgcn         &$0.039$       &$\underline{0.226}$                 &$0.036$   &$0.858$ &$0.192$   &$0.172$  &$0.134$\\ 
\gcnlpa       &$0.043$       &$0.116$                 &$0.023$   &$0.801$ &$0.167$   &$0.150$  &$0.075$\\
\hline
\end{tabular}
\label{tab: real_exres_ap}
\end{table}

We conclude that the current techniques are insufficient for solving the task of multi-label node classification based on the following observations.
\begin{itemize}[leftmargin=*]\setlength\itemsep{-0.5em}
    \item Simple baselines like \deepwalk which are almost never used for comparison with GNNs outperform GNNs on multiple real world datasets and almost all synthetic datasets.
    \item \hgcn which is designed for multi-class datasets with low homophily do not always show a considerable improvement over other methods even if the considered datasets have low label homophily.
    \item Techniques like \lanc which are designed specifically for multi-label datasets do not shown considerable improvements. 
\end{itemize}

\begin{table}[!h]
\setlength{\tabcolsep}{3pt}
\caption{Average Precision (mean) on the synthetic datasets. $r_{ori\_feat}$ and $r_{homo}$ refer to the fraction of original features and the label homophily, respectively.}
\centering
\small
\begin{tabular}{l|ccccc|ccccc}
\hline
\multirow{2}{*}{Method}        & \multicolumn{5}{c|}{$r_{ori\_feat}$}  & \multicolumn{5}{c}{$r_{homo}$}\\
      & $0.0$          &$0.2$          &$0.5$            &$0.8$           &$1.0$  & $0.2$               &$0.4$                &$0.6$                &$0.8$               &$1.0$\\
\hline 
\rule{0pt}{2.5ex}\mlp       &$0.172$ &$0.187$ &$0.220$ &$0.277$ &$0.343$ & $\bf{0.343}$  &$0.343$   &$0.343$   &$0.343$  &$0.343$\\ 
\deepwalk  &$\textbf{0.487}$ &$\textbf{0.487}$ &$\textbf{0.487}$ &$\bf{0.487}$ &$\bf{0.487}$  & $0.181$  &$\bf{0.522}$   &$\bf{0.813}$   &$\textbf{0.869}$ &$0.552$ \\ 
\lanc &$0.337$ &$0.342$ &$0.365$ &$0.353$ &$0.391$ &$0.190$ &$0.380$ &$0.434$ &$0.481$ &$\underline{0.629}$ \\
\gcn       &$0.313$ &$0.316$ &$0.311$ &$0.301$ &$0.337$ & $0.261$  &$0.343$   &$0.388$   &$0.450$ &$0.493$ \\ 
\gat       &$0.311$ &$0.339$ &$0.329$ &$0.338$ &$0.360$ & $0.172$  &$0.359$   &$0.390$   &$0.428$ &$0.439$ \\ 
\graphsage &$0.300$ &$0.328$ &$0.377$ &$0.393$ & $0.430$ & $0.289$  &$0.426$   &$0.458$   &$0.533$ &$0.553$ \\ 
\hgcn      &$\underline{0.376}$ &$\underline{0.401}$ &$\underline{0.427}$ &$\underline{0.442}$ &$\underline{0.467}$ & $\underline{0.297}$  &$\underline{0.484}$   &$\underline{0.512}$   &$0.572$ &$\textbf{0.652}$\\
\gcnlpa    &$0.337$ &$0.333$ &$0.368$ &$0.363$ &$0.391$ &$0.170$  &$0.408$   &$0.495$   &$\underline{0.604}$ &$0.583$\\
\hline
\end{tabular}
\label{tab: syn_exres}
\end{table}
\newpage
\impact{In the progress analysis of GNNs, multi-label node classification has been ignored. We fill in this gap with the analysis of the existing datasets and the proposal of the new datasets. We hope to encourage the community to look into this more realistic and general case of node classification. Our graph generator model and detailed data analysis can be useful in various applications having positive societal impact like protein function prediction using the multi-labelled protein interaction network.
}

\vskip 0.2in
\bibliography{sample}


\end{document}